\pdfoutput=1

\documentclass[11pt]{article}

\usepackage[]{acl}
\usepackage{booktabs}

\usepackage{times}
\usepackage{latexsym}
\usepackage{graphicx}

\usepackage[T1]{fontenc}

\usepackage[utf8]{inputenc}

\usepackage{microtype}

\title{The Dangers of Underclaiming:\\Reasons for Caution When Reporting How NLP Systems Fail}

\author{Samuel R. Bowman\\
New York University\\
\texttt{bowman@nyu.edu}}

\begin{document}
\maketitle
\begin{abstract}
Researchers in NLP often frame and discuss research results in ways that serve to deemphasize the field's successes, often in response to the field's widespread hype. Though well-meaning, this has yielded many misleading or false claims about the limits of our best technology. This is a problem, and it may be more serious than it looks: It harms our credibility in ways that can make it harder to mitigate present-day harms, like those involving biased systems for content moderation or resume screening. It also limits our ability to prepare for the potentially enormous impacts of more distant future advances. This paper urges researchers to be careful about these claims and suggests some research directions and communication strategies that will make it easier to avoid or rebut them.
\end{abstract}

\section{Introduction}

Over the last few years, natural language processing has seen a wave of surprising negative results overturning previously-reported success stories about what our models can do, and showing that widely-used models are surprisingly brittle \citep{jia-liang-2017-adversarial,niven-kao-2019-probing,mccoy-etal-2019-right}. This shows that many of our standard practices for evaluation and reporting can lead to unrealistically positive initial claims about what we can do. The resulting hype and overclaiming, whether intentional or not, are a problem. They can encourage the reckless deployment of NLP systems in high-stakes settings where they can do significant harm. They also threaten the health and credibility of NLP as a research field, and thereby threaten our ability to influence applied stakeholders or attract funding.

Fortunately, these results have led to a surge of research and writing that proposes more thorough and cautious practices for the evaluation of model ability \citep{ribeiro-etal-2020-beyond,gardner-etal-2020-evaluating,kiela-etal-2021-dynabench,bowman-dahl-2021-will}. While we have only a limited ability to control the public narrative taking place through industry PR and the media, there's reason to be hopeful that we researchers are getting much better at avoiding the worst forms of overconfidence about our systems.
Less fortunately, this pattern of disappointment seems to have led to many instances of pessimism about model performance that are ungrounded from real empirical results. This leaves room for the research community's consensus about our capabilities to fall short of our actual capabilities. 

I call this issue \textit{underclaiming}, for lack of a better term,\footnote{While \textit{overclaiming} generally refers to overstating the effectiveness of \textit{one's own} methods or ideas, the phenomenon that I call underclaiming often involves downplaying the effectiveness of preexisting methods or ideas.} and argue that it is more dangerous than it might seem. It risks our credibility and thereby limits our ability to influence stakeholders in cases where our current systems are doing real harm. It also limits our ability to accurately forecast and plan for the impacts that may result from the deployment of more capable systems in the future. If we can truly reach near-human-level performance on many of the core problems of NLP, we should expect enormous impacts which will be potentially catastrophic if not planned for.

\begin{figure}[t]
\setlength\fboxsep{0.04\columnwidth}
\centering
\noindent
\fbox{\parbox{0.91\columnwidth}{
    \includegraphics[width=0.91\columnwidth]{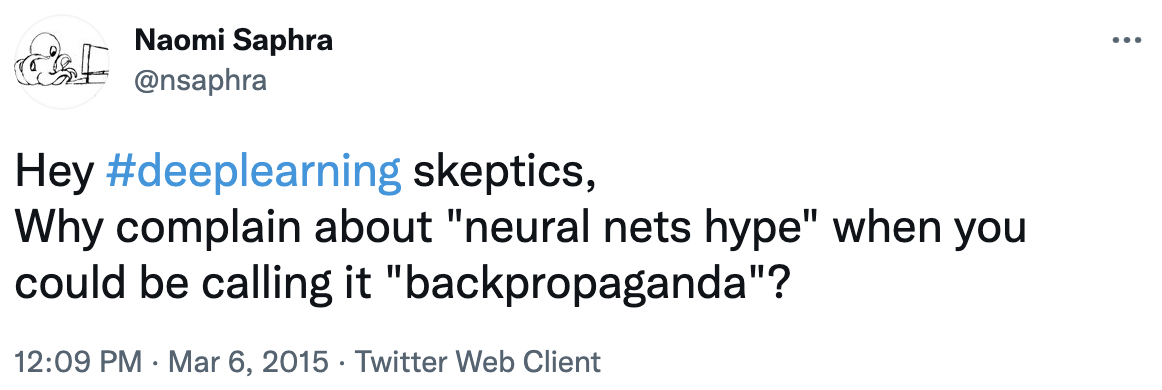}
}}
    \caption{Hype is a problem. The opposite of hype isn't necessarily better. (Quoted with permission.)}
    \label{fig:tweet}
\end{figure}

In this paper, I lay out case studies demonstrating four types of underclaiming, focusing especially on writing and citation practices. I then argue that it is a problem. I close by sketching some ways of reducing the prevalence of this kind of underclaiming, including straightforward best practices in writing and evaluation, a proposed rule of thumb for writing and reviewing, improvements to tooling for analysis and benchmarking, and research directions in model performance forecasting and test set design.

\section{Underclaiming: Case Studies}


\begin{figure}[t]
\setlength\fboxsep{0.04\columnwidth}
\centering
\noindent
\fbox{\parbox{0.88\columnwidth}{\centering
    \includegraphics[width=0.65\columnwidth]{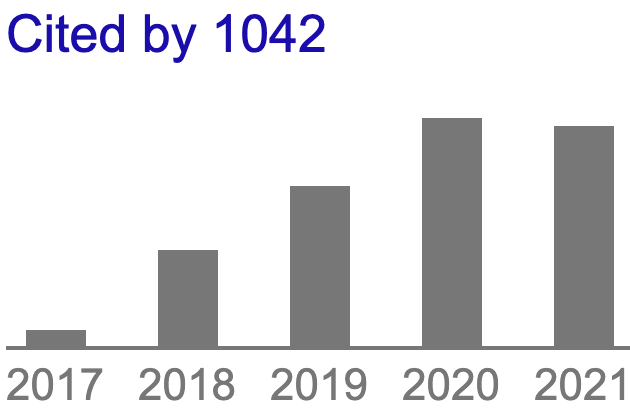}
}}
    \caption{\citet{jia-liang-2017-adversarial} remains widely cited according to Google Scholar. The original work pointed out major unexpected limitations in neural networks trained from scratch on the SQuAD reading comprehension task. However, many of these citing works use it to imply that modern pretrained systems---developed more recently than 2017---show these same limitations.}
\end{figure}

This paper addresses the phenomenon of scholarly claims that imply state-of-the-art systems are significantly less capable than they actually are. This takes on several forms, including misleading presentations of valid negative results from weak or dated baseline models, misleading claims about the limits of what is conceptually possible with machine learning, and misleading reporting of results on adversarially collected data. 


\subsection{Negative Results on Weaker Models}
\label{sec:dated}

Despite many surprises and setbacks, NLP research seems to have made genuine progress on many problems over the last few years. In light of this, discussions about the limitations of systems from past years don't straightforwardly apply to present systems. The first two cases that I present involve failures to contextualize claims about the failures of weaker past systems:

\begin{table}[t]
    \centering\small
    \begin{tabular}{lrrrr}
    \toprule
    Model & Year & SQuAD & AS & AOS \\
    \midrule
    ReasoNet Ensemble &2017 & 81 & 39 & 50  \\
    BERT-Base &2018 & 87 & 64 & 72 \\
    XLNet-Base&  2019 & \textbf{89} & \textbf{69} & \textbf{77} \\
    \bottomrule
    \end{tabular}
    \caption{F1 results on the original SQuAD development set and the two \citeauthor{jia-liang-2017-adversarial} adversarial evaluation sets. Results cover the best-performing SQuAD model studied by \citeauthor{jia-liang-2017-adversarial}---ReasoNet \citep{shen2017reasonet}---and the newer BERT and XLNet models \citep{devlin-etal-2019-bert,xlnet}, as tested by \citet{zhou-etal-2020-curse}.  While I am not aware of results from more recent models on the this data, progress through 2019 had already cut error rates in half.}
    \label{reasonnet}
\end{table}

\paragraph{Adversarial Examples for SQuAD} \citet{jia-liang-2017-adversarial} published one of the first demonstrations of serious brittleness in neural-network-based systems for NLU, showing that a simple algorithm could automatically augment examples from the SQuAD benchmark \citep{rajpurkar-etal-2016-squad} in a way that fool many state-of-the-art systems, but not humans. This work prompted a wave of much-needed analysis and a corresponding lowering of expectations about the effectiveness of neural network methods.

However, the results in \citeauthor{jia-liang-2017-adversarial} predate the development of modern pretraining methods in NLP \citep{peters-etal-2018-deep,radford2018improving,devlin-etal-2019-bert}, and the best systems studied in this work have more than twice the error rate of the current state of the art. While I am not aware of any results from current state-of-the-art systems on this data, results from 2019 systems suggest that we are making substantial progress (Table \ref{reasonnet}). We have no reason to expect, then, that the failures documented in this work are quantitatively or qualitatively similar to the failures of current systems.

However, papers that cite these results often present them with no discussion of the model under study, yielding misleading implications. For example, the award-winning work of \citet{linzen-2020-accelerate} cites the \citeauthor{jia-liang-2017-adversarial} result to justify this claim:
\begin{quote}
    [F]or current deep learning systems: when tested on cases sampled from a distribution that differs from the one they were trained on, their behavior is unpredictable and inconsistent with that of humans
\end{quote}
The chief concern in this context is the claim that this failure applies to \textit{current deep learning systems} in general, and the corresponding unjustified implication that these failures are a fundamental or defining feature of neural network language models. Looking only to highly-cited works from the last two years that cite \citeauthor{jia-liang-2017-adversarial}, similar statements can be found in \citet{xu2020adversarial}, \citet{zhang2020semantics}, and others. 

\paragraph{The Long Shadow of BERT}

While the case of \citeauthor{jia-liang-2017-adversarial} is especially striking since it deals with models that predate pretraining entirely, a similar effect is much more common in a subtler form: Most \textit{analysis} papers that identify limitations of a system come out well after the system description paper that claims the initial (typically positive) results.  BERT, first released in fall 2018, has been a major locus for this kind of analysis work, and continues to be long after its release.
Looking to a random sample of ten papers from the NAACL 2021 analysis track that study pretrained models,\footnote{Papers studying only BERT: \citet{white-etal-2021-non,slobodkin-etal-2021-mediators,bian-etal-2021-attention,cao-etal-2021-low,pezeshkpour-etal-2021-empirical}. Papers studying other models predating fall 2019: \citet{wallace-etal-2021-concealed,hall-maudslay-cotterell-2021-syntactic,hollenstein-etal-2021-multilingual,bitton-etal-2021-automatic,du-etal-2021-towards}} none of them analyze models that have come out since summer 2019, and five only study BERT, representing a median lag of nearly three years from the release of a model to the publication of the relevant analysis.\footnote{A similar analysis of the late-2021 EMNLP conference, conducted after peer review for the present paper, shows a slightly better median lag of two years.}

This analysis work is often valuable and these long timelines can be justifiable: Good analysis work takes time, and researchers doing analysis work often have an incentive to focus on older models to ensure that they can reproduce previously observed effects. Even so, this three-year lag makes it easy to seriously misjudge our progress.

In particular, this trend has consequences for the conclusions that one would draw from a broad review of the recent literature on some problem: A review of that literature will contrast the successes of the best current systems against the weaknesses of the best systems \textit{from an earlier period}. In many cases, these weaknesses will be so severe as to challenge the credibility of the successes if they are not properly recognized as belonging to different model generations.

The BERT-only results, though, represent a clear missed opportunity: There exist newer models like RoBERTa and DeBERTa \citep{liu2019roberta,he2020deberta} which follow nearly identical APIs and architectures to BERT, such that it should generally be possible to reuse any BERT-oriented analysis method on these newer models without modification. In many cases, these newer models are different enough in their performance that we should expect analyzing them to yield very different conclusions: For example, BERT performs slightly \textit{worse} than chance on the few-shot Winograd Schema commonsense reasoning test set in SuperGLUE \citep{levesque2011winograd,wang2019superglue}, while DeBERTa reaches a near-perfect 96\% accuracy. How much better would our understanding of current technology be if a few of these works had additionally reported results with DeBERTa?


\subsection{Strong Claims about \textit{Understanding}}

The influential work of \citet{bender-koller-2020-climbing} is centered on the claim that:
\begin{quote}
[T]he language modeling task, because it only uses form as training data, cannot in principle lead to learning of meaning.
\end{quote}
The proof of this claim is straightforward and convincing under some (but not all) mainstream definitions of the word \textit{meaning} in the context of NLP: If \textit{meaning} deals with the relationship between language and some external nonlinguistic reality, then a system that can only ever interact with the world through language cannot access meaning. 

This argument does not, on its own, make any prediction about the behavior of these models on tasks that take place entirely through the medium of language.
Under this definition, a translation system is acting \textit{without reference to meaning} even if it has a rich, structured internal model of the world, and even it interprets sentences with reference to that model when translating: As long as that model of the world is developed solely using language, no \textit{meaning} is involved.\footnote{See \citet{merrill2021meaning} for some limits on how closely such a model can correspond to the real world and \citet[\S2.6.3]{bommasani2021opportunities} for further discussion of the implications of \citeauthor{bender-koller-2020-climbing}'s arguments for NLP.}

In addition, this argument does not justify any strong prediction about the behavior of models which are trained primarily, but \textit{not exclusively}, on a language modeling objective, as with models that are fine-tuned to produce non-textual outputs like labels, or models which are trained in a multimodal language-and-vision regime. 

While this core claim is sound and important, public discussion of the paper has often repeated the claim in ways that imply stronger conclusions about model behavior. \citet{utama-etal-2020-mind}, for example, write
\begin{quote}
Researchers have recently
studied more closely the success of large fine-tuned LMs in many NLU tasks and found that models are simply better in leveraging biased patterns instead of capturing a better notion of language understanding for the intended task \citep{bender-koller-2020-climbing}.
\end{quote}, misleadingly suggesting that this result deals with the outward performance of specific language models on tasks.

In another vein, \citet{jang2021noier} make the straightforward claim that 

\begin{quote}
Bender and Koller (2020) show that it is impossible to learn the meaning of language by only leveraging the form of sentences.
\end{quote}
but they then use that claim to motivate a new regularization technique for language models, which does nothing to change the fact that they are trained on form alone. In this context, it is hard to avoid the incorrect inference that \citeauthor{bender-koller-2020-climbing} show a \textit{specific and contingent} problem with recent language models---which could be mitigated by better regularization.

Similar claims can be found in many other citing works \citep{utama-etal-2020-mind,van-noord-etal-2020-character,hovy-yang-2021-importance,sayers:hal-03230287,peti2021croatian,jang2021noier}.
While \citeauthor{bender-koller-2020-climbing} raise important points for discussion, these strong implications in citing works are misleading and  potentially harmful.

\subsection{Adversarially Collected Test Sets}

\textit{Adversarially collected} test sets \citep{bartolo-etal-2020-beat,nie-etal-2020-adversarial,kiela-etal-2021-dynabench}---or test sets composed of examples that some target system gets wrong---have recently become a popular tool in the evaluation of NLP systems. Datasets of this kind are crowdsourced in a setting where an example-writer can interact with a model (or ensemble) in real time and is asked to come up with examples on which the model fails. Writers are generally  incentivized to find these failure cases, and the test section(s) of the resulting dataset will generally consist \textit{exclusively} of such cases.

This process produces difficult test sets and it can be a useful tool in understanding the limits of existing training sets and models \citep{williams2020anlizing}. However, the constraint that a specified system \textit{must} fail on the test examples makes it difficult to infer much from absolute measures of test-set performance: As long as a model makes \textit{any errors at all} on \textit{any} possible inputs, then we expect it to be possible to construct an adversarial test set against the model, and we expect the model to achieve zero test accuracy on that test set. We can further infer that any models that are \textit{sufficiently similar} to the adversary should also perform very poorly on this test set, regardless of their ability. Neither of these observations would tell us anything non-trivial about the actual abilities of the models.

What's more, in many NLU data collection efforts, a large share of annotator disagreements represent subjective judgments rather than clear-cut errors \citep{pavlick-kwiatkowski-2019-inherent}. This means that even a perfectly careful and perfectly well-qualified human annotator should be expected to disagree with the majority judgment on some examples, and will thereby be coded as having made errors. It is, therefore, possible to create an adversarial test set for which a careful human annotator would achieve 0\% accuracy. Absolute performance numbers on adversarially-collected test sets are meaningless as measures of model capabilities.

Adversarially-collected test sets are often used in standard experimental paradigms, and these caveats about the interpretation of results are not always clear when numbers are presented. Sampling papers that cite \citet{nie-etal-2020-adversarial}, for example, it is easy to find references that do not mention the adversarial design of the data and that therefore make claims that are hard to justify:\footnote{I focus here about claims about the \textit{absolute} performance level of models. Whether adversarially collected test sets are appropriate for comparing the \textit{relative} effectiveness of models is a largely orthogonal issue \citep{bowman-dahl-2021-will,kaushik2021efficacy,phang2021adversarially}.}
\citet{talmor-etal-2020-olmpics} use the results from \citeauthor{nie-etal-2020-adversarial} to claim that ``LMs do not take into account the presence of negation in sentences'', and
\citet{hidey2020deseption} use them to justify the claim that ``examples for numerical reasoning and lexical inference have been shown to be difficult.'' \citet{10.1145/3442188.3445922} misleadingly describe a form of adversarial data collection\footnote{AFLite \citep{Bras2020AdversarialFO} uses  ensembles of \textit{weak} models to filter data. This avoids the most direct \textit{0\% accuracy} concerns, but it can still provide arbitrarily large distortions to absolute  performance in a way that is disconnected from any information about the skill or task that a dataset is meant to test.} as a method for the ``careful manipulation of the test data to remove spurious cues the systems
are leveraging'', and cite results on such data to argue that ``no actual language understanding is taking place in
LM-driven approaches''.
\citet{liu2020adversarial} similarly use absolute results on the adversary models to back up the trivial but easily-misread claim that BERT-style models ``may still suffer catastrophic failures in adversarial scenarios.'' 



\section{A Word on Hype}

The previous section has laid out some ways in which the mainstream NLP research community makes unjustifiable claims about the limitations of state-of-the-art methods. These claims do not make the opposite phenomenon, \textit{hype}, any less real or any less harmful.
While hype is likely most severe in industry PR and in the media,\footnote{Consider the 2017 Huffington Post headline ``Facebook Shuts Down AI Robot After It Creates Its Own Language.''}
it is nonetheless still prevalent in the research literature. In one especially clear example, a prominent paper claiming of human parity in machine translation performance \citep{hassan2018achieving} severely overstates what has been accomplished relative to commonsense intuitions about what a human-level translation system would do  \citep{toral-etal-2018-attaining,laubli-etal-2018-machine,zhang-toral-2019-effect,graham-etal-2020-statistical}. 

I do not aim to argue that overclaiming or hype is acceptable or safe. 
Combating hype should be fully compatible with the goals laid out in this paper, and broad-based efforts to improve our practices in evaluation, analysis, writing, and forecasting should help reduce both underclaiming and hype.

\section{Why Underclaiming is Harmful}


Research papers are generally most useful when they're true and informative. A research field that allows misleading claims to go unchallenged is likely to waste its time solving problems that it doesn't actually have, and is likely to lose credibility with serious funders, reporters, and industry stakeholders. This is the most obvious reason that we should be concerned about underclaiming, but it is not the whole story.
This loss of insight and credibility can seriously challenge our ability to anticipate, understand, and manage the impacts of deploying NLP systems. This is especially true of impacts that are contingent on NLP technologies \textit{actually working well}, which we should expect will become more substantial as time goes on.

\subsection{Present-Day Impact Mitigation} 

The deployment of modern NLP systems has had significant positive and negative impacts on the world. Researchers in NLP have an ethical obligation to inform (and if necessary, pressure) stakeholders about how to avoid or mitigate the negative impacts while  realizing the positive ones. 
Most prominently, typical applied NLP models show serious biases with respect to legally protected attributes like race and gender \citep{bolukbasi2016man,rudinger-etal-2018-gender,Parrish2021BBQAH}. 
We have no reliable mechanisms to mitigate these biases and no reason to believe that they will be satisfactorily resolved with larger scale. 
Worse, it is not clear that even superhuman levels of fairness on some measures would be satisfactory: Fairness norms can conflict with one another, and in some cases, a machine decision-maker will be given more trust and deference than a human decision-maker would in the same situation \citep[see, e.g.,][]{Rudin2020Age,fazelpour2020algorithmic}.
We thus are standing on shaky moral grounds when we deploy present systems in high-impact settings, but they are being widely deployed anyway \citep[e.g.][]{amazon,bertsearch,fb}. Beyond bias, similar present-day concerns can be seen around issues involving minority languages and dialects, deceptive design, and the concentration of power \citep[\S3.3]{joshi-etal-2020-state,10.1145/3442188.3445922,kenton2021alignment}.

Persuading the operators of deployed systems to take these issues seriously, and to mitigate harms or scale back deployments when necessary, will be difficult. Intuitively, researchers concerned about these harms may find it appealing to emphasize the limitations of models in the hope that this will discourage the deployment of harmful systems. This kind of strategic underclaiming can easily backfire: Models are often both useful and harmful, especially when the operator of the system is not the one being harmed. If the operator of some deployed system sees firsthand that a system is effective for their purposes, they have little reason to trust researchers who argue that that same system \textit{does not understand language}, or who argue something similarly broad and negative. They will then be unlikely to listen to those researchers' further claims that such a system is harmful, even if those further claims are accurate. 

\subsection{Preparing for Future Risks}

We can reasonably expect NLP systems to improve over the coming decades. Even if intellectual progress from research were to slow, the dropping price of compute should allow us to continue to reap the benefits of larger-scale training \citep{kaplan2020scaling,brown2020language}. This improvement in capabilities is likely to amplify both the harms and benefits of language technology.

We have good reason to expect that this further progress in NLP, over many years or decades, will lead to upheavals in areas like education, medicine, law, and the service sector more broadly, as well as making mass surveillance and misinformation campaigns far more effective and opening up additional new use cases that will be hard for us to foresee \citep{brundage2018malicious,tamkin2021understanding,bommasani2021opportunities}. One can reasonably expect that the positive and negative impacts of these upheavals will far exceed the impacts that our technologies have produced to date. In turn, NLP researchers who want to ensure that their career has a net-positive impact on the world should be concerned with these possibilities.

How does this relate to underclaiming? It will be difficult to do the necessary technical, social, and governance work to prepare for these advances if we do not have a clear picture of our current capabilities, and it will be difficult to convince outside stakeholders to act appropriately to mitigate these risks if we don't acknowledge that we have made, and are making, real progress toward effective language technology.

Looking somewhat further into the future, a substantial community of philosophers, economists, and general ML researchers are concerned that highly-capable AI systems---of the kind that could plausibly be developed through existing ML research paradigms---are extremely dangerous by default \citep{bostrom2012superintelligent,critch2020ai,christian2020alignment,ord2020precipice,russelbook}. Expert forecasts suggest that this could take place within a few decades \citep{grace2018will}. If these hypotheses hold, and if we are poorly prepared for these developments, the worst-case outcomes could be catastrophic, even threatening the existence of human civilization on some views. 

Investments in research into these potential catastrophic risks from advanced machine learning have become substantial: Funding from one 
foundation alone has totaled over \$200M USD.\footnote{\url{https://www.openphilanthropy.org/giving/grants}} Concerns about risks from AI have also been the stated motivation for a significant fraction of the work from DeepMind and OpenAI, which both have access to even greater amounts of funding. The British Prime Minister Boris Johnson recently made a speech calling for further investment on the floor of the UN General Assembly \cite{boris}.

Spurred on in particular by the shift in emergent capabilities from GPT-2 to GPT-3, the attention of these AI risk researchers has also been increasingly centered on language models and similar self-supervised multimodal models \citep[\S4.9]{irving2018ai,stiennon,hendrycks2020aligning,kenton2021alignment,wu2021recursively,bommasani2021opportunities}.
Despite the scale of this research, and its recent shift of focus toward language models, there has been little interaction between the research communities working on long-term AI risk and on NLP.

The facts that AI risk research is growing in influence and that it is increasingly focused on language models put NLP in an exceptionally strange and troubling situation as a field.
To the extent that these concerns are valid, they represent an urgent call for reprioritization within NLP research to favor safety-relevant areas like interpretability, control, and evaluation over scaling, and to push for better oversight and regulation of large-scale research \citep{dafoe2018ai}:
Even a small risk of a globally significant catastrophe warrants a dramatic response. 
On the other hand, to the extent that these concerns are unfounded or are built on misunderstandings about the possible trajectories of ML research, it would be quite valuable to correct this misunderstanding. Correcting the record could redirect these resources and, more significantly, reduce the risk that popular or regulatory pressure will snuff out the positive potential of NLP technologies. 

\section{Catastrophic Risks}

Because these more speculative concerns around advanced artificial intelligence are rarely discussed in the NLP literature, I will here offer a brief overview of that work. Recent writing tends to focus on four clusters of hypotheses:

\paragraph{Unaccountable Organizations} Highly-capable AI is likely to lead to highly-profitable applications, making the institutions that first develop it quite powerful. It is also likely to be able to displace human labor in technical fields to a large extent, increasing the relative value of capital over labor, and making it easier for the leaders of these organizations to take unpopular actions unilaterally. In the longer term, highly-capable AI may also  contribute to the effectiveness of persuasion campaigns, further insulating these organizations from outside pressure. These forces could conspire to make the companies or  governments that first produce highly-capable AI almost entirely unaccountable, and allowing their decisions to play a major role in the trajectory of humanity as a whole \citep{ord2020precipice}.

\paragraph{Alignment and Robustness Failures} Even if a system is deployed by an actor with good intentions and substantial oversight, good outcomes are not guaranteed.
As AI systems become more capable, they become capable of effecting---directly or indirectly---significant force on the outside world.
In these cases, it becomes crucial that they behave in ways that we would endorse, even when they are pushed into unfamiliar new situations. This requires both that the systems be optimized for the right objectives  and that the systems actually internalize and generalize those objectives correctly.

Specifying and using \textit{safe} objectives, such that aggressively optimizing them does not produce catastrophic outcomes, is difficult \citep{critch2020ai}. Human preferences are complex, making the problem of specifying an objective that rules out unintended bad behavior non-trivial.
Goodhart's law\footnote{in the formulation of \citet{strathern1997improving}: ``When a measure becomes a target, it ceases to be a good measure.''} means that many objectives that serve as good proxies for what we want in in familiar situations can break down in new situations. 

Further, training large models with high precision is difficult. A small flaw in a highly-capable system's learned understanding of its objective can cause catastrophic failures, even if the true intended objective would have been safe \citep{hubinger2019risks}.

\begin{figure}
    \centering
    \includegraphics[width=0.9\columnwidth]{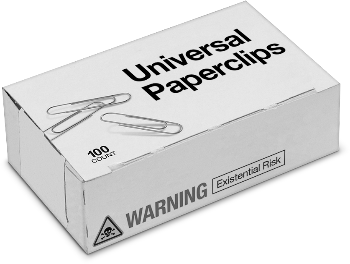}
    \caption{Downplaying the capabilities of current ML systems makes it less likely that we'll be well prepared for the impacts that come from developing highly-capable future sytsems. That can be bad. Image from \citet{lantz2017paperclips}.\label{clippy}}
\end{figure}

\paragraph{Instrumentally-Convergent Subgoals} The \textit{instrumental convergence} hypothesis holds that systems that are optimizing for benign objectives, once they become sufficiently capable, have a predictable reason to take on dangerous \textit{subgoals}---like accumulating large amounts of computational, economic, or political power---to maximize the odds that their primary objectives are achieved \citep{bostrom2003ethical,omohundro2008basic,bostrom2012superintelligent}.\footnote{This is exemplified by the thought experiment of the \textit{paperclip maximizer} (Figure \ref{clippy}), which points out that a machine tasked with manufacturing as many paperclips as possible, if sufficiently capable, should be expected to turn nearly all matter on earth into paperclips. While this vision of a single system acting alone on such a trivial objective is unrealistic, it demonstrates the key hypothesis that almost any reasonable-sounding goal starts to conflict with basic human needs if a sufficiently capable system pursues it single-mindedly.} 
Even with merely near-human-like levels of performance, the ability of computational models to be copied and accelerated gives them considerable leeway to act in un-human-like ways.
Systems that interact with humans only through text, or systems whose goals are circumscribed to a well-defined task like question answering, are not exempt from this concern \citep{armstrong2012thinking}.

\paragraph{Risks Will Be Difficult to Spot} Human-level capabilities are likely to emerge first from large machine learning models that, like modern neural networks, are not directly interpretable. This means that it may be difficult to spot ways in which a model is unsafe or to forecast ways in which its behavior might change in novel settings \citep{critch2020ai}. 

Further, we should expect highly-capable AI systems to be \textit{useful} in the short term, giving potential users a strong incentive to deploy them as soon as they are affordable, even if their safety is not guaranteed. This means that it is not enough that it simply be \textit{possible} for us to develop safe systems, it is additionally necessary that it be nearly as easy and nearly as affordable as developing unsafe systems \citep{irving2018ai}.

\paragraph{So What?}

None of these arguments is conclusive in its current form, but as far as I am aware, all have resisted straightforward attempts at falsification. All four are potentially applicable to neural network-based models and to models which operate primarily through language.
While the nascent field of \textit{AI alignment} has proposed some mechanisms by which we might mitigate these risks, work in this area is still largely exploratory, with no clear research agenda in place to ensure that powerful models will be safe \citep{hadfield2016cooperative,irving2018ai,critch2020ai,kenton2021alignment,am2021general}. 
If these arguments hold, significant further work is needed to avoid catastrophe. This will be difficult to achieve without a clear accounting of the abilities and limitations of current and plausible near-future systems. In particular, we will need enough foresight to be able to see substantial progress of this kind coming well in advance, to avoid the complacency that comes with the perception that worrying about impacts from powerful AI is like worrying about ``overpopulation on mars'' \citep[quoting Andrew Ng]{mars}.

\section{Ways to Do Better}

The core issue in this paper is one of sloppy communication about results. The most straightforward step that we can take to remedy underclaiming is to simply use the same practices that we already use to avoid overclaiming: The peer-review process already polices overclaiming to a significant extent, and most researchers have learned to be careful about overclaiming in their writing. We should apply high standards of evidence to our own empirical claims and those of others, both in peer-reviewed venues and in more informal scientific communication, even when those claims are negative and cloaked in a frame of individual or field-level modesty.

Beyond this, there are specific best practices or research directions that can help make these mistakes harder to make:

\paragraph{A Rule of Thumb} In light of the issues with negative results on older models discussed in Section \ref{sec:dated}, it could be productive to introduce a new  heuristic when reviewing or evaluating papers that discuss model failures.\footnote{While a corresponding rule could be helpful in the context of results describing the \textit{success} of a machine learning system on some evaluation, the asymmetry here is intentional: Successes are likely to be deliberately replicated from one generation of models to the next, while the opposite is true of failures.} In the spirit of the Bender Rule \citep{bender2019rule}, 
I propose: 

\begin{quote}
    
When describing the failure of a machine learning model on some empirical evaluation, make it clear
\begin{itemize}
    \item[i.] what kind of model has failed,
    \item[ii.] whether the model is significantly less capable than the current state of the art in the domain, and
    \item[iii.] whether the evaluation was deliberately set up to trick that model or another model like it.
\end{itemize}

\end{quote}


\paragraph{Better Evaluation}

The pervasiveness of underclaiming can likely be attributed in part to the ineffectiveness of current evaluation practices in many areas of NLP. When impressive numbers on widely-used benchmarks are usually followed by disappointment, suggesting that good evaluation numbers don't translate to effective systems, it is rational to treat new encouraging results with extreme skepticism.

Better benchmarks and evaluation practices could help mitigate this by providing a firmer ground on which to make positive claims about system capacities.\footnote{Though \citet{raji2021ai} point out ways in which better benchmarking \textit{alone} is unlikely to be fully satisfactory.} In practice, research into more effective crowdsourcing and benchmark design 
and research into better statistical reporting and publication norms \citep{dodge-etal-2019-show,card-etal-2020-little,rogers-augenstein-2020-improve,van-miltenburg-etal-2021-preregistering}
seem especially high-impact under this lens. 

\paragraph{Better Analysis}

We can help address the time-lag issue discussed in Section \ref{sec:dated} by building tooling to make it easier to adapt existing analysis techniques to new models seamlessly. Leaderboards that integrate conventional benchmarking with analysis can be especially helpful by making this largely automatic \citep{wang-etal-2018-glue,dua2019orb,gehrmann2021gem,ma2021dynaboard}. More broadly, careful analysis work, targeted at broadly understanding the capacities of capable models, will be valuable in helping to forecast and mitigate the worst risks from future systems \citep{elhage2021circuits,Ganguli2022PredictabilityAS}.

\paragraph{Better Forecasting} 

\textit{Scaling laws} results in NLP \citep{hestness2017deep,kaplan2020scaling,brown2020language,zhang-etal-2021-need}
offer the promise that we can predict the performance of \textit{future} larger-scale machine learning models on at least some metrics. This line of work is still nascent, and successes to date have largely focused on loss values rather than more interpretable measures of capability. Further developing these methods, as well as others that allow us to better forecast near future progress, should be helpful. Better forecasting will provide a useful way to sanity-check future claims \citep{dvigna2019predict} and will help improve the responsiveness of model analysis by enabling us to prepare analysis methods and datasets that \textit{anticipate} future capabilities.

\section{Additional Related Work}

While much of this paper discusses the state of the NLP literature, a few related works warrant emphasis as starting points for further reading:

\citet{bender-koller-2020-climbing}, \citet{10.1145/3442188.3445922}, and \citet{raji2021ai} discuss the role of hype in driving bad outcomes from the development of language technology. \citet{jin-etal-2021-good} and \citet{rogers-2021-changing} offer broader discussion of how to ensure that the net impact of near-future NLP deployments on the world is positive. \citet{morris-etal-2020-reevaluating} and \citet{hauser2021bert} highlight overly strong negative claims in papers analyzing models' robustness to synonym substitution. 

Looking to the longer term,  \citet[\S4.9]{bommasani2021opportunities} provides an introduction to the AI risk and AI alignment literature from a perspective that emphasizes NLP and language. 
\citet{welty2019metrology}, \citet{linzen-2020-accelerate}, \citet{ribeiro-etal-2020-beyond}, \citet{raji2021ai}, \citet{bowman-dahl-2021-will}, and \citet{dehghani2021benchmark}, among many others, discuss the challenges involved in designing evaluations that yield trustworthy and accurate depictions of the capabilities of ML models.


\section{Conclusion}

Like many research fields that have a tight connection to technological practice, 
NLP has long struggled to avoid inflated expectations about the capabilities of state-of-the-art tools. This remains a serious issue. However, this paper argues that our attempts to avoid hype often overshoot: Instead of merely correcting overly optimistic claims about our capabilities, we replace them with overly \textit{pessimistic} claims.

Making misleading claims is generally a bad sign for the health and credibility of a scientific field, and the stakes are high: NLP technologies are implicated in a range of serious real-world harms, and plausible future elaborations of these technologies are potentially much more dangerous still. Our ability to mitigate existing harms will depend on our ability to make reliably credible claims about the limitations of our systems. Our ability to mitigate future harms will depend on our ability to accurately anticipate, recognize, agree upon, and report upon emerging capabilities. Both of these goals are seriously hampered by claims that current technologies are less capable than they in fact are.


Better evaluation, better tooling for model analysis, and better mechanisms for technical forecasting should all contribute to making these pessimistic claims easier to avoid or debunk. However, this problem is ultimately one of scientific communication, and to solve it fully, we will need to use the tools and norms of science to better police false or misleading claims. The stakes are high.

\section*{Acknowledgements}

The ideas and arguments in this work were developed through conversation (live or on Twitter) with more researchers than I can name, including Emiel van Miltenburg, Deb Raji, Paul Christiano, Geoffrey Irving, Rohin Shah, Jacob Steinhardt, Catherine Olsson, Nick Beckstead, Alex Tamkin, Daniel Dewey, Alex Ray, and Robin Jia, as well as audiences at UT Austin, Georgia Tech, UPenn, University College London, CMU, Bar Ilan University, Tel Aviv University, Technion, Hebrew University, Unbabel, Instituto Superior T\'ecnico, UChicago, and TTI-C, and contributors to the AI Alignment Forum. Anna Rogers, Owain Evans, Alex Wang, Jacob Steinhardt, Jason Phang, Jared Kaplan, Alex Cristia, Tal Linzen, Jonathan Uesato, and four anonymous ARR reviewers provided feedback on drafts. Any errors or dubious rhetoric are my own.

This project has benefited from financial support by Eric and Wendy Schmidt (made by recommendation of the Schmidt Futures program) and Apple. This material is based upon work supported by the National Science Foundation under Grant Nos. 1922658 and 2046556. Any opinions, findings, and conclusions or recommendations expressed in this material are those of the author(s) and do not necessarily reflect the views of the National Science Foundation. 

\bibliography{anthology}
\bibliographystyle{acl_natbib}

\appendix

\end{document}